\documentclass[letterpaper]{article} %
\usepackage{aaai2026}  %
\usepackage{times}  %
\usepackage{helvet}  %
\usepackage{courier}  %
\usepackage[hyphens]{url}  %
\usepackage{graphicx} %
\usepackage{natbib}  %
\usepackage{caption} %
\usepackage{algorithm}
\usepackage{algorithmic}

\usepackage{newfloat}
\usepackage{listings}
\DeclareCaptionStyle{ruled}{labelfont=normalfont,labelsep=colon,strut=off} %
\floatstyle{ruled}
\newfloat{listing}{tb}{lst}{}
\floatname{listing}{Listing}

\usepackage{xcolor} 
\usepackage{multirow}
\usepackage{subcaption}
\usepackage{amsmath}
\usepackage{tabularx}
\newcolumntype{C}{>{\centering\arraybackslash}X}
\usepackage{booktabs}
\usepackage{algorithm}
\usepackage{algorithmic}
\usepackage{placeins}

\usepackage{listings}
\DeclareCaptionStyle{ruled}{labelfont=normalfont,labelsep=colon,strut=off} %
\floatstyle{ruled}
\newfloat{listing}{tb}{lst}{}
\floatname{listing}{Listing}
\title{GRIM: Task-Oriented Grasping with Conditioning on Generative Examples}

\author {
    Shailesh\textsuperscript{\rm 1},\\
    Alok Raj\textsuperscript{\rm 1},
    Nayan Kumar\textsuperscript{\rm 1}, 
    Priya Shukla\textsuperscript{\rm 2},\\
    Andrew Melnik\textsuperscript{\rm 3},
    Michael Beetz\textsuperscript{\rm 3},
    Gora Chand Nandi\textsuperscript{\rm 2}
}
\affiliations {
    \textsuperscript{\rm 1}IIT Dhanbad, India\\
    \textsuperscript{\rm 2}IIIT Allahabad, India\\
    \textsuperscript{\rm 3}University of Bremen, Germany
}

\begin{document}

\maketitle
\begin{figure*}[t]
    \includegraphics[width=\textwidth]{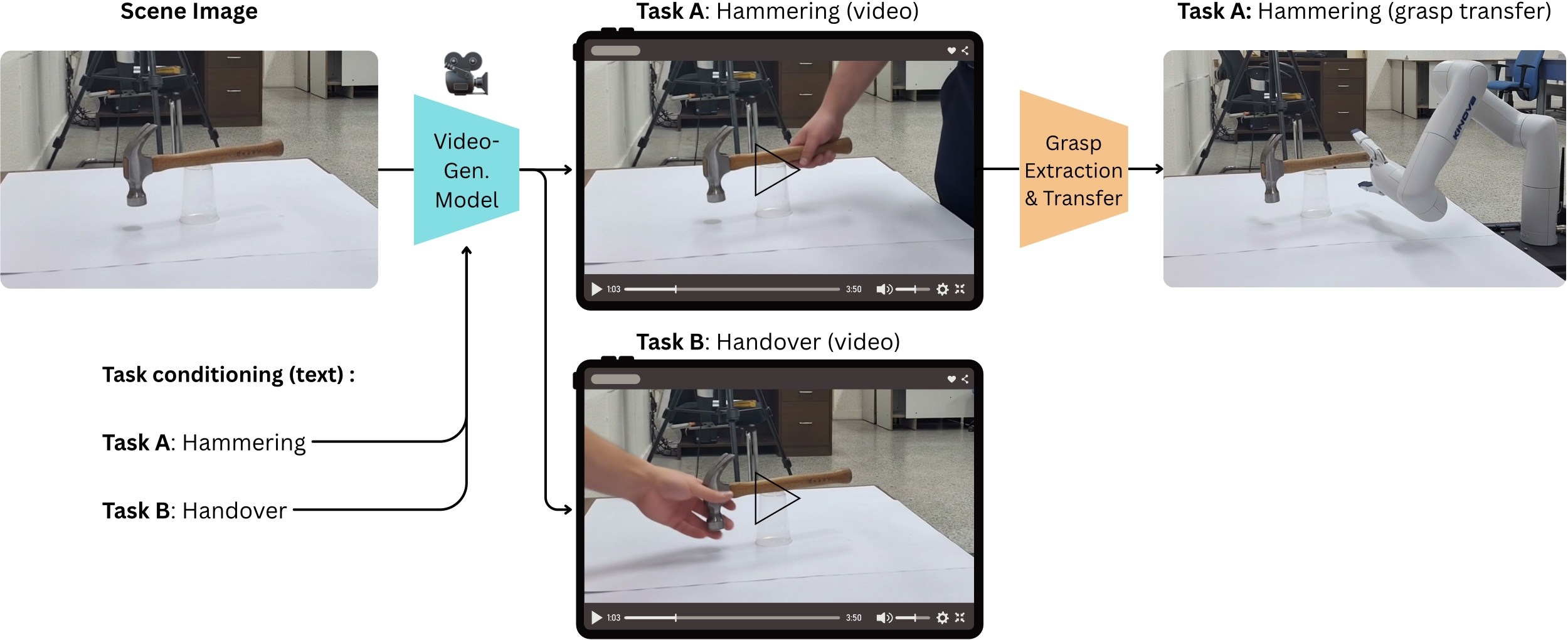}
    \caption{The GRIM framework for task-oriented grasp synthesis. From a single scene image, the VGM generates task-specific video examples, such as hammering (Task A) and handover (Task B). Grasps are extracted from these generated videos and then transferred to a robotic arm to execute the specified task in the real world as shown for hammering (Task A).}
    \label{fig:phase0}
\end{figure*}
\begin{abstract}

Task-Oriented Grasping (TOG) requires robots to select grasps that are functionally appropriate for a specified task — a challenge that demands an understanding of task semantics, object affordances, and functional constraints. We present GRIM (Grasp Re-alignment via Iterative Matching), a training-free framework that addresses these challenges by leveraging Video Generation Models (VGMs) together with a retrieve–align–transfer pipeline. Beyond leveraging VGMs, GRIM can construct a memory of object–task exemplars sourced from web images, human demonstrations, or generative models. The retrieved task-oriented grasp is then transferred and refined by evaluating it against a set of geometrically stable candidate grasps to ensure both functional suitability and physical feasibility. GRIM demonstrates strong generalization and achieves state-of-the-art performance on standard TOG benchmarks.

\end{abstract}
\begin{links}
    \link{Code, Dataset \& Appendix}{https://grim-tog.github.io/}
\end{links}

\begin{figure*}[t]
    \centering
    \includegraphics[width=\textwidth]{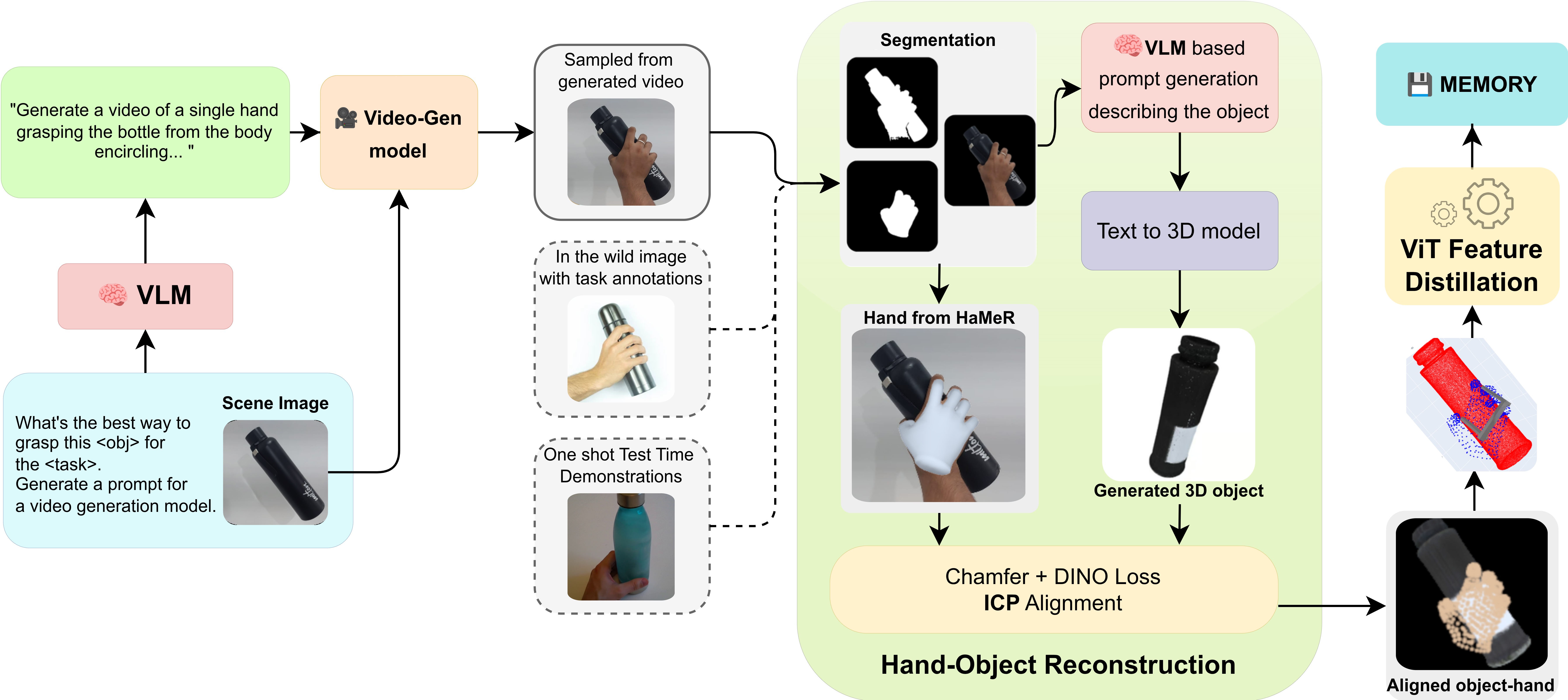}
    \caption{Our memory creation pipeline. A diverse set of inputs (AI-generated video frames, web images, human demonstrations) are processed by a hand-object reconstruction module \citep{wu2024reconstructing}. This yields an object mesh and a corresponding task-oriented grasp pose. We enrich the object mesh with dense DINO features to create a feature mesh, which is stored in memory alongside the task label and grasp pose.}
    \label{fig:phase1}
\end{figure*}

\section{Introduction}

The ability for robots to physically interact with the world is fundamental to their utility. While grasp synthesis has made significant strides in achieving geometric stability, true manipulation intelligence lies in selecting grasps that are functionally suitable for a specific goal. This problem, known as Task-Oriented Grasping (TOG), moves beyond the question of ``Can I pick this up?'' to ``How should I pick this up to complete task X?''. For example, a hammer must be grasped by its handle to be used for hammering, not by its head. This requires a deep understanding of object affordances, task semantics, and the functional constraints they impose. A primary bottleneck for progress in TOG is the data-scarcity problem. Supervised learning methods \citep{murali2020taskgrasp, tang2023graspgpt} are powerful, but depend on large, manually annotated datasets that specify which grasps are suitable for which tasks. %

To address these challenges, we propose \textbf{GRIM} (Grasp Re-alignment via Iterative Matching), a novel \textbf{training-free} framework that leverages the power of pre-trained foundation models in a retrieve-align-transfer pipeline \citep{kuang2024ram, dipalo2024dinobot}. Our approach follows the pipeline shown in Figure \ref{fig:phase0}. Since GRIM leverage grasping from generated video and the underlying VGMs so far too computationally heavy for real-time execution, we also built a memory option \citep{melnik2018world} for grasp retrieval. GRIM can build an extendable memory (Figure \ref{fig:phase1}) of object-task interactions from diverse and easily accessible sources: in-the-wild images from the web, on-demand human demonstrations, or synthetic data from generative models. 

\begin{figure*}[htbp]
    \centering
    \includegraphics[width=\textwidth]{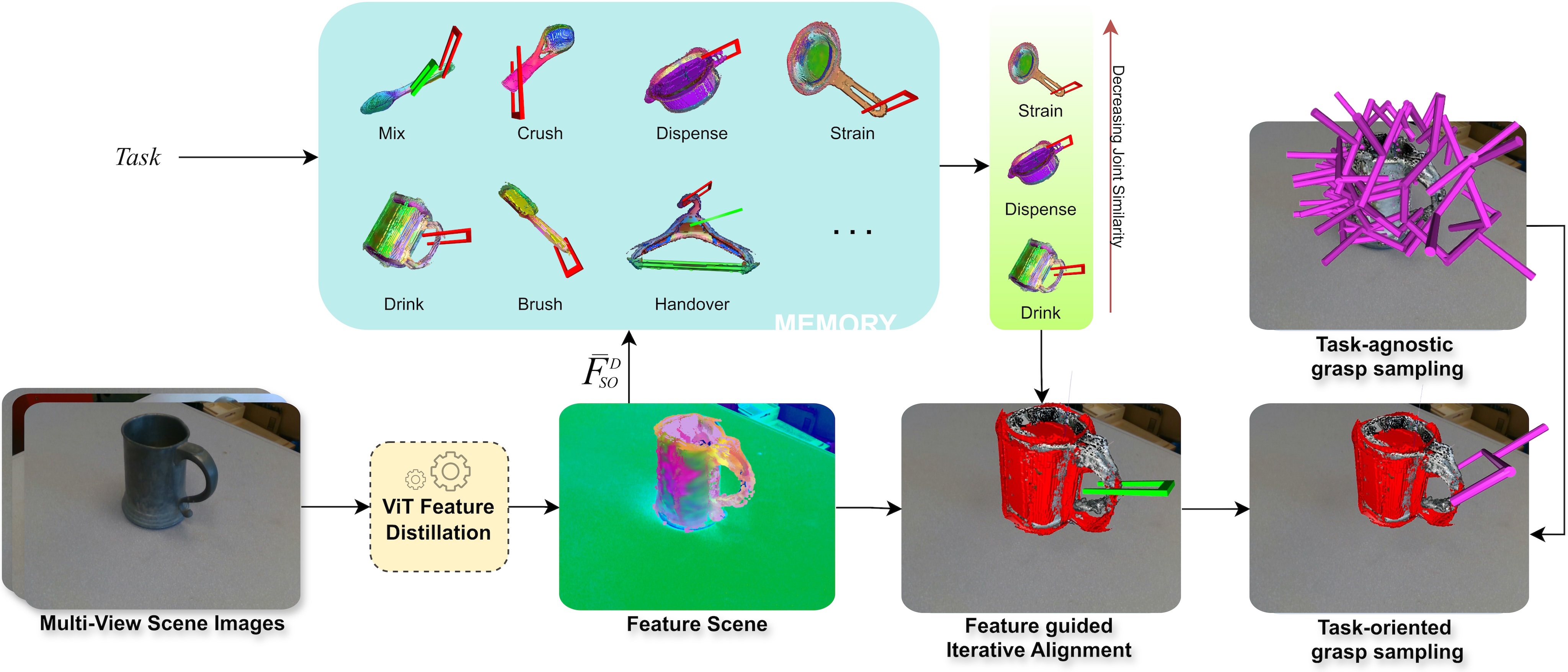}
    \caption{The GRIM pipeline for a given scene object and task. \textbf{(1) Retrieval:} The system queries its memory using joint visual and task similarity to find the best matching prior experience (a cup for the task `drink`). \textbf{(2) Alignment:} The retrieved memory object (red point cloud) is aligned with the scene object (grey point cloud) using our feature-guided iterative alignment. The colors on the objects represent PCA-reduced DINO features, showing semantic correspondence. \textbf{(3) Transfer \& Refine:} The grasp from the memory object is transferred to the scene object and used to select the best among a set of task-agnostic, stable grasp candidates (cluster of purple grasps), resulting in the final task-oriented grasp (single purple grasp).}
    \label{fig:phase2}
\end{figure*}

GRIM's memory retrial workflow is as follows (Figure \ref{fig:phase2}):
\begin{enumerate}
    \item \textbf{Retrieve:} It queries its memory to find the most relevant prior experience, using a joint similarity metric that considers both the visual appearance of the object (via DINO embeddings \citep{oquab2024dinov2learningrobustvisual}) and the semantics of the task description (via CLIP embeddings \citep{radford2021learningtransferablevisualmodels}).
    \item \textbf{Align:} It robustly aligns the 3D point cloud of the retrieved memory object with the scene object. This is a key contribution, employing a coarse-to-fine strategy that first uses PCA-reduced DINO features for a semantically-aware coarse alignment, followed by a precise ICP \citep{Besl1992AMF} refinement.
    \item \textbf{Transfer \& Refine:} The task-specific grasp pose from the memory instance is transferred to the aligned scene object. This transferred pose then serves as a powerful prior to select and refine the best grasp from a set of pre-computed, geometrically stable candidates for the scene object.
\end{enumerate}

Our main contributions are:
\begin{itemize}
    \item a training-free framework that leverages VGMs together with a retrieve–align–transfer pipeline for TOG which demonstrates generalization to both novel objects and novel tasks
    \item A flexible and scalable memory construction pipeline that integrates object-task experiences from diverse sources, including a novel application of generative AI, circumventing the need for manually annotated datasets.
    \item A novel 3D alignment strategy that prioritizes semantic correspondence over geometric shape. By matching dense DINO features, our method works effectively even with sparse, partial point clouds where traditional geometry-based alignment techniques often fail.
\end{itemize}

\section{Related Work}

Task-Oriented Grasping (TOG) research has evolved from analytical methods to data-driven techniques, with a recent shift towards leveraging large-scale pre-trained models.

\subsection{Data-Driven Approaches}
Early data-driven methods learned direct mappings from object classes and tasks to grasps \citep{6385563, Liu2019CAGECG}. However, these approaches often lacked semantic understanding and struggled to generalize \citep{tang2023graspgpt}. To inject semantic knowledge, subsequent works utilized knowledge bases (KBs) and probabilistic logic \citep{5649406, huang2022inner, LIU2023104294, Ardn2019LearningGA, 10.1007/978-3-319-13413-0_5}, but these systems often require significant engineering to construct and scale the KBs. %

The release of the TaskGrasp dataset by \citet{murali2020taskgrasp} was a significant step, enabling methods like GCNGrasp which uses a Graph Convolutional Network. However, such methods are inherently limited by the contents of their training data and knowledge graph, struggling to generalize to concepts unseen during training. More recent works like GraspGPT \citep{tang2023graspgpt} and GraspMolmo \citep{deshpande2025graspmolmo} leverage Large Language Models (LLMs) and Vision-Language Models (VLMs) to incorporate open-world knowledge, improving generalization \cite{mikami2024natural}. Nevertheless, these models still rely on a foundational training phase on task-specific datasets \citep{Tang2023TaskOrientedGP, Jin2024ReasoningGV, Nguyen2023LanguageConditionedAD}, inheriting the associated data acquisition bottleneck. %

GRIM diverges fundamentally from these paradigms. It is entirely training-free, obviating the need for task-specific grasp annotations. By dynamically building a memory from heterogeneous data, it directly tackles the data scarcity and annotation challenges that constrain prior methods.

\subsection{Training-Free and Retrieval-Based Approaches}
The advent of powerful foundation models has spurred the development of training-free TOG methods. Many approaches use LLMs or VLMs to provide semantic guidance, mapping a language command to a region on an object where a grasp should be executed \citep{lerftogo2023, mirjalili2024langraspusinglargelanguage, 10801661}. %
While these methods avoid training, they typically only provide coarse spatial priors (e.g., ``grasp the handle"), not directly executable 6D grasp poses.

Closer to our work are retrieval-based methods. RTAGrasp \citep{dong2024rtagrasp} also proposes a training-free approach using a memory of human demonstrations. It retrieves a relevant video and uses 2D feature matching to transfer a grasp point. While effective, its reliance on 2D matching can be ambiguous and less robust to viewpoint changes. RoboABC \citep{ju2024roboabc} uses CLIP to retrieve contact points but struggles to determine the full 6D grasp pose, particularly the crucial grasp orientation.

GRIM builds upon the strengths of retrieval but makes several key improvements. Our retrieval is guided by a joint 3D visual (DINO \citep{oquab2024dinov2learningrobustvisual}) and task-semantic (CLIP \citep{radford2021learningtransferablevisualmodels}) similarity. %
Crucially, we introduce a robust, semantically-aware 3D alignment strategy that aligns entire object point clouds, not just 2D features \citep{dipalo2024dinobot}. %
This allows for a more precise transfer of the full 6D grasp pose, which is then further refined against the scene object's specific geometry. This holistic process addresses both "where" and "how" to grasp with high precision and adaptability, without the limitations of pre-defined datasets or explicit training.

\section{Methodology}

We introduce GRIM (Grasp Re-alignment via Iterative Matching), a training-free framework for TOG. Our approach follows a retrieve-align-transfer pipeline, detailed below.

\subsection{Memory Creation}
To generalize to novel scenes, we construct a memory $\mathcal{M}$ of object-task experiences from diverse data sources. Each instance in $\mathcal{M}$ is a tuple $(F_M, G_t, T, O)$, containing the object's feature mesh $F_M$, a 6D task-oriented grasp pose $G_t$, the corresponding task description $T$, and the object name $O$.

The pipeline to create a single memory instance (Figure \ref{fig:phase1}) begins with an image or video frame $I_{HO}$ depicting a functional grasp. We use a hand-object reconstruction model \citep{wu2024reconstructing} to extract the object mesh and hand mesh. We then derive a 6D parallel-jaw gripper pose $G_t$ from the hand mesh. This conversion is done by first identifying the centroids of hand segments: the thumb, the combined index and middle fingers, and the palm. The gripper's center (translation) is defined as the midpoint between the centroid of the thumb and the combined centroid of the opposing fingers. The vector connecting these centroids establishes the closing direction, and the palm's centroid provides a reference point to determine the approach vector.

To create the feature mesh $F_M$, we sample points from the object mesh and compute a dense DINOv2 feature vector for each point, effectively creating a semantic descriptor field on the object's surface, similar to \citet{wang2023d3fields} and \citet{arjun2024splatr}.

While GRIM primarily learns from AI-generated videos, our pipeline is flexible and can incorporate additional data sources as well:
\subsubsection{AI-Generated Videos:} To create a scalable and diverse data source, we leverage generative AI \citep{melnik2024video}.
For an object and task from a source like TaskGrasp \citep{murali2020taskgrasp}, we prompt a VLM (Gemini Pro) to generate a detailed textual description of a video showing the correct grasp. This description, along with a starting image frame, is then used as a prompt for a VGM such as VEO2 \cite{GoogleVeo2}, Grok Imagine, etc. to generate a short video. We sample a frame from this video to serve as $I_{HO}$. This process allows for cheap, large-scale creation of functionally-grounded grasp data.
\subsubsection{In-the-Wild Web Images:} We use images scraped from the web that show human grasping actions. For each image, we use a VLM to generate a plausible task description $T$.
\subsubsection{Test-Time Expert Demonstrations:} Our framework supports lifelong learning. If the robot fails on a task, a human can provide a single-image demonstration, which is seamlessly processed and added to the memory $\mathcal{M}$, improving future performance on similar tasks \citep{malato2024zero}. 

\subsection{Memory Retrieval}
Given a novel scene containing a target object (represented by its point cloud $P_{SO}$ with per-point DINO features $F_{SO}^D$) and a task command $T_S$, we retrieve the most relevant memory instance.

First, we compute a global visual descriptor $\bar{F}_{SO}^D$ for the scene object by averaging its per-point DINO features. We encode the task command $T_S$ into a text embedding $E_{T_S}$ using CLIP's text encoder.

For each memory instance $i \in \mathcal{M}$ with its global object descriptor $\bar{F}_{MO,i}^D$ and task embedding $E_{T_{M,i}}$, we compute a joint similarity score:
\begin{equation} \label{eq:joint_similarity}
S_{\text{joint}}(i) = \alpha \cdot \text{sim}_{\text{cos}}(\bar{F}_{SO}^D, \bar{F}_{MO,i}^D) + (1-\alpha) \cdot \text{sim}_{\text{cos}}(E_{T_S}, E_{T_{M,i}})
\end{equation}
where $\text{sim}_{\text{cos}}(\cdot, \cdot)$ is the cosine similarity and $\alpha$ is a hyperparameter balancing visual and task similarity (we use $\alpha=0.5$). The memory instance $(F_M^*, G_t^*, T^*, O^*)$ with the highest $S_{\text{joint}}$ is selected for the next stage.

\subsection{Semantic 3D Alignment}
After retrieving a memory object (source point cloud $P_{MO}$, DINO features $F_{MO}^D$), we must align it to the scene object (target point cloud $P_{SO}$, features $F_{SO}^D$). A purely geometric alignment like standard ICP would fail if the objects have different shapes (e.g., aligning a metal spatula to a plastic one). We therefore propose a coarse-to-fine alignment strategy guided by semantic features.

\textbf{Coarse Alignment:} To reduce the dimensionality and noise of the DINO features, we apply PCA, projecting both $F_{MO}^D$ and $F_{SO}^D$ into a lower 4-dimensional space ($D_{\text{PCA}}=4$). We then perform a grid search over a discretized set of initial rotations to find a promising coarse alignment. Specifically, we sample 8 steps for each of the three Euler angles (roll, pitch, yaw), resulting in $8^3 = 512$ candidate rotations $\{R_i\}$. For each candidate, we compute a transformation $T_{\text{init},i}$ that aligns the point cloud centroids and applies the rotation. The quality of this initial transformation is evaluated using a joint feature-geometric score. For each point in the transformed source cloud, we find its $K=3$ nearest neighbors in the target cloud and compute a cost based on a weighted sum of the squared Euclidean distance ($w_g=10$) and the feature dissimilarity ($w_f=100$). By heavily weighting the feature component, we prioritize finding a semantically meaningful alignment over a purely geometric one. The top 10 transformations with the lowest cost are selected as candidates for the fine refinement stage.

\textbf{Fine Refinement:} The best coarse alignment is then used to initialize the Iterative Closest Point (ICP) algorithm. This standard ICP step refines the alignment to be geometrically precise. This two-step process, where semantics guide the initial guess and geometry refines it, allows for robust alignment even between objects that are semantically similar but geometrically distinct. The final output is a transformation $T_{\text{final}}$ that maps points from the memory object's coordinate frame to the scene object's frame.

\subsection{Grasp Transfer and Refinement}
With the alignment $T_{\text{final}}$, we transfer the task-oriented grasp $G_M$ from memory to the scene object: $G_S = T_{\text{final}} \cdot G_M$. However, due to small alignment errors or geometric differences, $G_S$ may not be perfectly stable or executable.

To find an optimal, executable pose, we follow a sample-and-refine strategy inspired by \citet{dong2024rtagrasp}. First, we use a task-agnostic grasp sampler, AnyGrasp \citep{fang2023anygrasprobustefficientgrasp}, to generate a set of $N$ geometrically stable grasp candidates $\{G_{A,i}\}_{i=1}^N$ on the scene object, each with a geometric quality score $S_{\text{geo},i}$.

We then re-rank these candidates based on their compatibility with our transferred task-oriented grasp $G_S = (R_S, \mathbf{t}_S)$. We define a task-compatibility score $S_{\text{task},i}$ for each candidate grasp $G_{A,i} = (R_{A,i}, \mathbf{t}_{A,i})$:
\begin{equation} \label{eq:stask}
S_{\text{task},i} = \underbrace{(\mathbf{v}_{\text{target}} \cdot \mathbf{v}_{A,i})}_\text{Orientation Sim.} + \underbrace{\exp\left(-\frac{\|\mathbf{t}_{A,i} - \mathbf{t}_{S}\|^2}{2\sigma^2}\right)}_\text{Position Sim.}
\end{equation}
where $\mathbf{v}_{\text{target}}$ and $\mathbf{v}_{A,i}$ are the approach vectors of the grasps (e.g., the z-axis of the gripper frame), and $\sigma$ is a bandwidth parameter (set to 0.02m). This score rewards candidates that are close in both position and orientation to the transferred task-centric pose.

The final score for each candidate is a weighted sum of its task compatibility and geometric quality:
\begin{equation} \label{eq:sfinal}
S_i = w_{\text{task}} S_{\text{task},i} + w_{\text{geo}} S_{\text{geo},i}
\end{equation}
We heavily prioritize task-compatibility by setting $w_{\text{task}}=0.95$ and $w_{\text{geo}}=0.05$, as AnyGrasp already ensures candidates have high geometric quality. The grasp candidate $G_{A}^*$ with the highest final score $S_i$ is selected for execution.

\begin{table*}[t]
    \centering
    \caption{Per-category Average Precision on novel object instances from the TaskGrasp dataset.}
    \label{tab:my_wide_label}
    \begin{tabularx}{\linewidth}{@{} l *{9}{C} @{}}
        \toprule
        \multirow{2}{*}{\textbf{Method}} & \multicolumn{9}{c}{Novel Instances} \\
        \cmidrule(lr){2-10}
        & Paint roller & Brush & Tongs & Strainer & Frying Pan & Fork & Mortar & Ice Scrapper & Pizza Cutter \\
        \midrule
        Random     & 0.30 & 0.66 & 0.23 & 0.24 & 0.32 & 0.26 & 0.31 & 0.60 & 0.50 \\
        RTAGrasp   & 0.39 & \textbf{0.93} & 0.28 & 0.55 & 0.42 & 0.35 & 0.37 & \textbf{0.91} & 0.57 \\
        GraspMolmo   & 0.56 & 0.73  & 0.55 & 0.44 & 0.46 & 0.53 & 0.66 & 0.65 & 0.58 \\
        \midrule
        GRIM(Ours) & \textbf{0.89} & 0.90 & \textbf{0.58} & \textbf{0.58} & \textbf{0.60} & \textbf{0.76} & \textbf{0.72} & 0.71 & \textbf{0.92} \\
        \bottomrule
    \end{tabularx}
\end{table*}
\section{Experiments and Results}

We conduct extensive experiments to evaluate GRIM's performance, focusing on its ability to generalize to novel objects and tasks.

\subsection{Experimental Setup}
\textbf{Baselines:} We compare GRIM against three representative baselines:
\begin{itemize}
    \item \textbf{Random:} A task-agnostic baseline that randomly selects a geometrically stable grasp from the candidates provided by AnyGrasp.
    \item \textbf{RTAGrasp \citep{dong2024rtagrasp}:} A state-of-the-art training-free method that uses 2D feature matching to transfer grasps from a video memory.
    \item \textbf{    GraspMolmo \citep{deshpande2025graspmolmo}:} A state-of-the-art learning-based VLM, which was fine-tuned on a mixture of its primary synthetic dataset (PRISM, 379k examples) and a portion of the TaskGrasp.
\end{itemize}

\textbf{Dataset:} We evaluate all methods on the TaskGrasp dataset \citep{murali2020taskgrasp}, which provides object point clouds and annotated task-oriented grasps. To rigorously test generalization, we use two challenging splits:
\begin{itemize}
    \item \textbf{Held-out Objects:} The memory contains no objects of the same category as the test object.
    \item \textbf{Held-out Tasks:} The memory contains no examples of the task being performed, even if it has seen the object category before.
\end{itemize}

\textbf{Memory:} Our memory for GRIM and RTAGrasp is constructed from a combination of 180 AI-generated video frames, 15 web images, and 15 human demonstrations, totaling 210 instances. This small size highlights the data efficiency of our approach. To ensure a fair comparison with RTAGrasp, we build its memory from the same source images and derive its required 2D grasp points from our 6D poses.

\textbf{Evaluation Metric:} Following standard practice, we evaluate the methods on their ability to identify the correct task-oriented grasps from a set of proposals. We use the 25 annotated grasps for each object instance in TaskGrasp as candidates. A predicted grasp is considered correct if it is one of the positive examples for the given task. We report the Mean Average Precision (mAP) over all object-task pairs. Since, these 25 grasp poses doesn't have any geometric quality score, we set $w_{\text{geo}}=0.0$ for final score calculation.

\subsection{Quantitative Results}

GRIM's effectiveness and data efficiency are demonstrated in our quantitative evaluations (Table \ref{tab:ap_results}). On the full TaskGrasp dataset, GRIM achieves a Mean Average Precision (mAP) of 0.67. This result not only surpasses the state-of-the-art training-free method, RTAGrasp (0.58), but also, remarkably, outperforms GraspMolmo (0.62). This comparison is particularly significant: GraspMolmo is a powerful VLM trained on a massive dataset of 379,000 synthetic task-oriented grasp examples, whereas GRIM's memory contains only 210 instances from heterogeneous, un-curated sources. This result strongly validates our central thesis: by effectively retrieving and re-aligning functional priors from a small but diverse memory, it is possible to achieve superior generalization without relying on vast, expensive, and potentially biased training datasets. Additionally, Table \ref{tab:my_wide_label} shows a more granular insight for a few object categories from the dataset.

Furthermore, GRIM's performance advantage is most pronounced in the challenging generalization splits. On held-out objects and tasks, GRIM's mAP degrades by only ~3\%, whereas RTAGrasp's performance drops by over 10\%. This underscores the robustness of our 3D semantic alignment strategy, which successfully transfers functional knowledge even without direct categorical or task precedents—a scenario where 2D feature matching proves less effective.

To understand why our method works well, we tested it without its key parts in an ablation study (Table \ref{tab:ablation_study}). The results clearly show that semantic alignment is the most critical component. Without it (GRIM w/o Semantic Alignment), performance drops to 0.50 mAP, which is nearly as poor as the random baseline. This confirms that using features is crucial for aligning objects for a task, especially when their shapes differ. The grasp refinement step is also important. Without it (GRIM w/o Grasp Refinement), performance falls to 0.59 mAP. This means the transferred grasp is a good starting point for the task, but it must be fine-tuned to the scene object's geometry to be successful. In summary, both components are vital: semantic alignment provides the correct functional idea, and refinement makes that idea physically work.

A qualitative analysis further illuminates the behavior of the semantic alignment module, particularly its failure modes. Its performance is intrinsically linked to the fidelity of the input point cloud. In scenarios with severe sensor noise or extreme sparsity, the process of establishing dense feature correspondences can break down. This corrupts geometric priors like the centroid and leads to a flawed coarse alignment from which the local ICP refinement cannot recover. The final transferred grasp is consequently misplaced and functionally irrelevant. This underscores a key dependency: while GRIM is robust to partial views, its ability to reason functionally is contingent on receiving a partial point cloud of sufficient quality to support the crucial semantic alignment stage.

\begin{table}[htbp]
    \centering
    \caption{Mean Average Precision (mAP) on the TaskGrasp dataset. GRIM consistently outperforms all baselines, with particularly strong performance on the held-out splits, demonstrating superior generalization.}
    \label{tab:ap_results}
    \setlength{\tabcolsep}{5pt}
    \begin{tabular}{lccc}
        \toprule
        \textbf{Method} & \textbf{All Data} & \textbf{Held-out Obj.} & \textbf{Held-out Tasks} \\
        \midrule
        Random          & 0.49          & 0.41            & 0.43           \\
        RTAGrasp        & 0.58          & 0.52           & 0.51           \\
        GraspMolmo      & 0.62          & 0.57           & 0.55           \\
        \midrule
        \textbf{GRIM (Ours)}  & \textbf{0.67} & \textbf{0.65}  & \textbf{0.64}  \\
        \bottomrule
    \end{tabular}
\end{table}

\begin{table}[htbp]
    \centering
    \caption{Ablation study of GRIM's key components. Results are reported as Mean Average Precision (mAP) on the \textbf{full TaskGrasp dataset}, demonstrating the critical role of both semantic alignment and grasp refinement.}
    \label{tab:ablation_study}
    \begin{tabular}{lc}
        \toprule
        \textbf{Configuration} & \textbf{mAP (All Data)} \\
        \midrule
        \textit{Ablations:} \\
        GRIM w/o Semantic Alignment & 0.50 \\
        GRIM w/o Grasp Refinement & 0.59 \\
        \midrule
        \textbf{GRIM (Full Model)} & \textbf{0.67} \\
        \bottomrule
    \end{tabular}
\end{table}

\subsection{Real-World Robot Validation}
\begin{figure}[htbp]
    \centering
    \begin{subfigure}[t]{0.49\linewidth}
        \includegraphics[width=\textwidth]{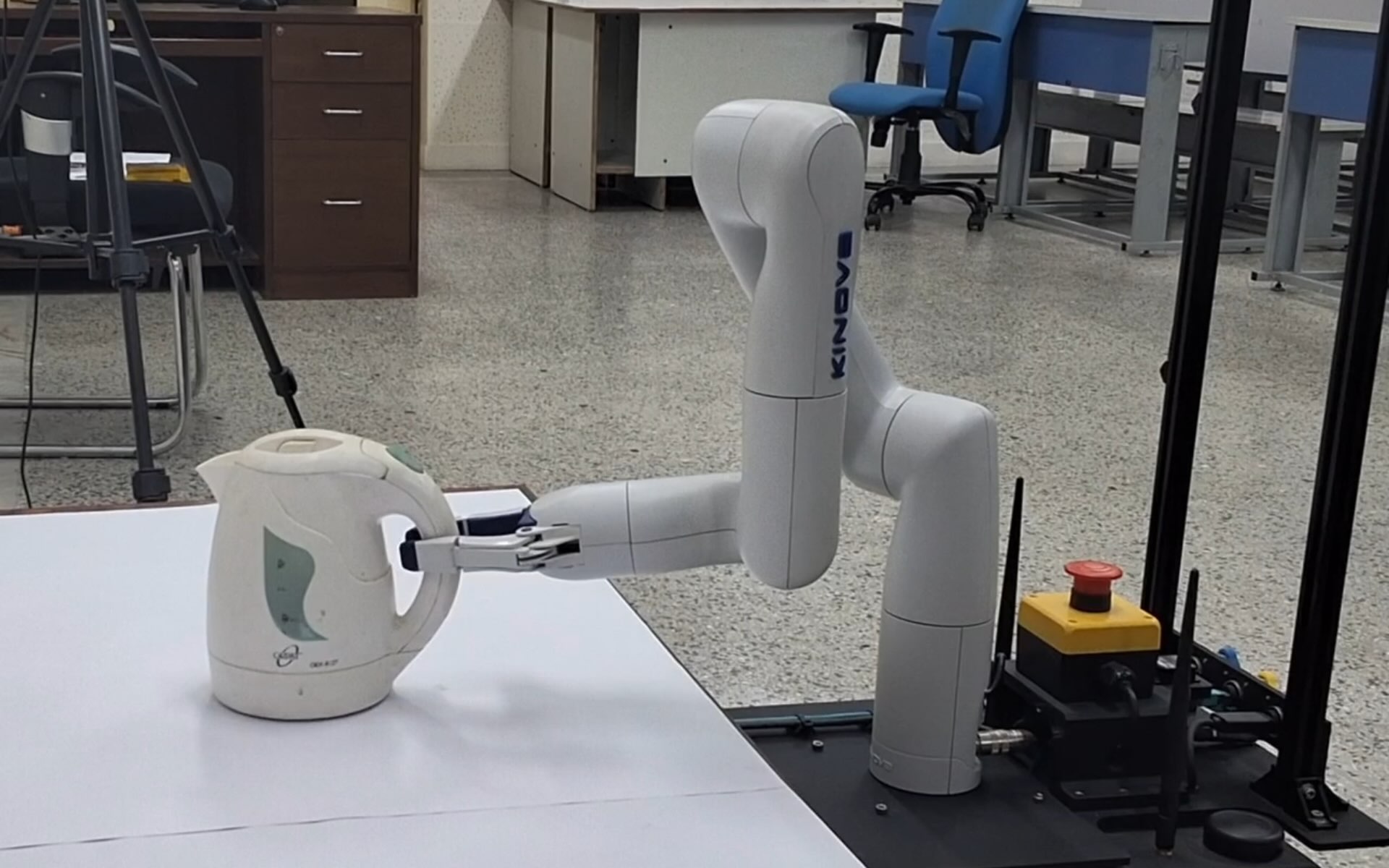}
        \caption{} %
        \label{fig:real_robot_mallet_plan}
    \end{subfigure}
    \begin{subfigure}[t]{0.49\linewidth}
        \includegraphics[width=\textwidth]{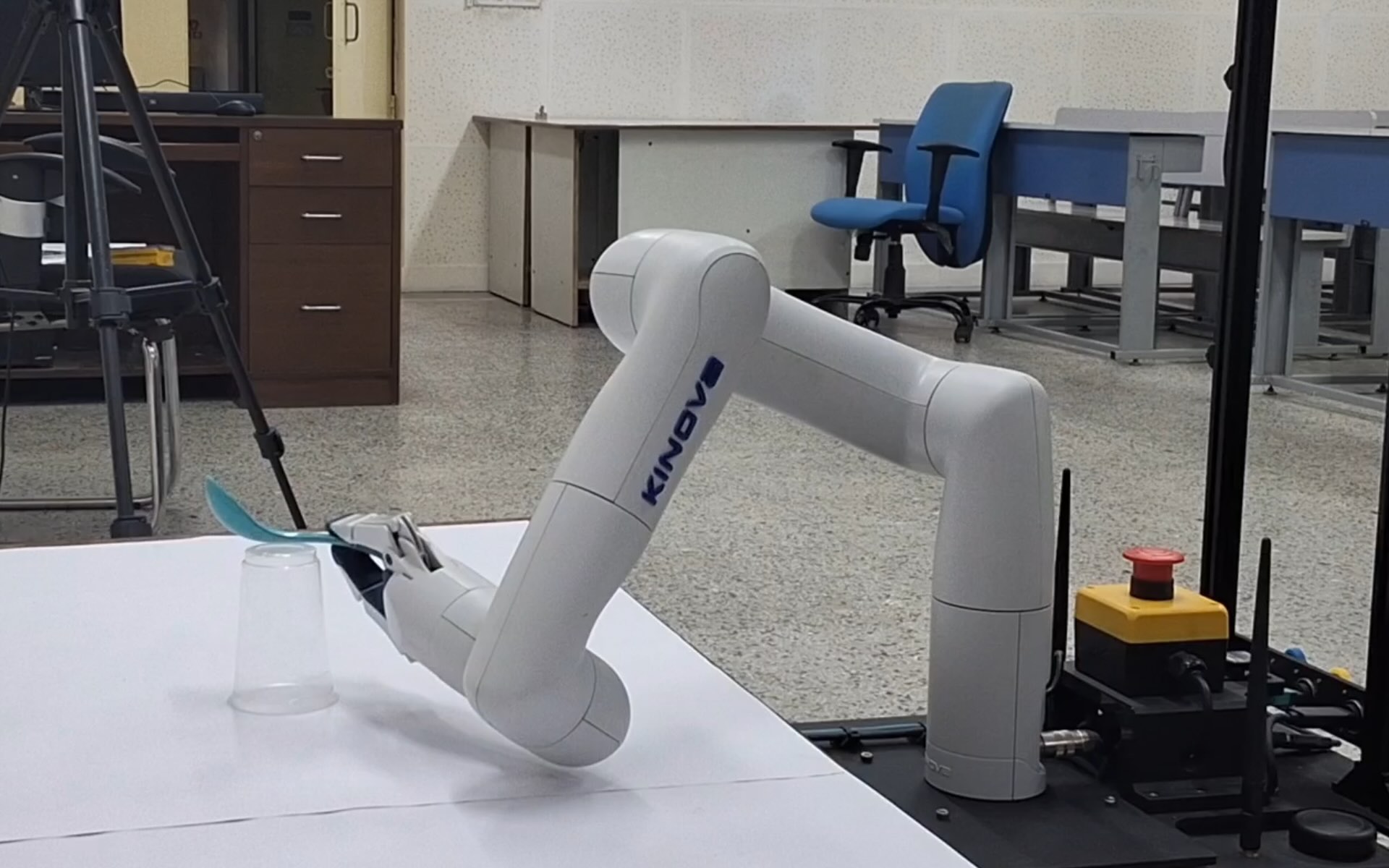}
        \caption{} %
        \label{fig:real_robot_mallet_exec}
    \end{subfigure}

    \begin{subfigure}[t]{0.49\linewidth}
        \includegraphics[width=\textwidth]{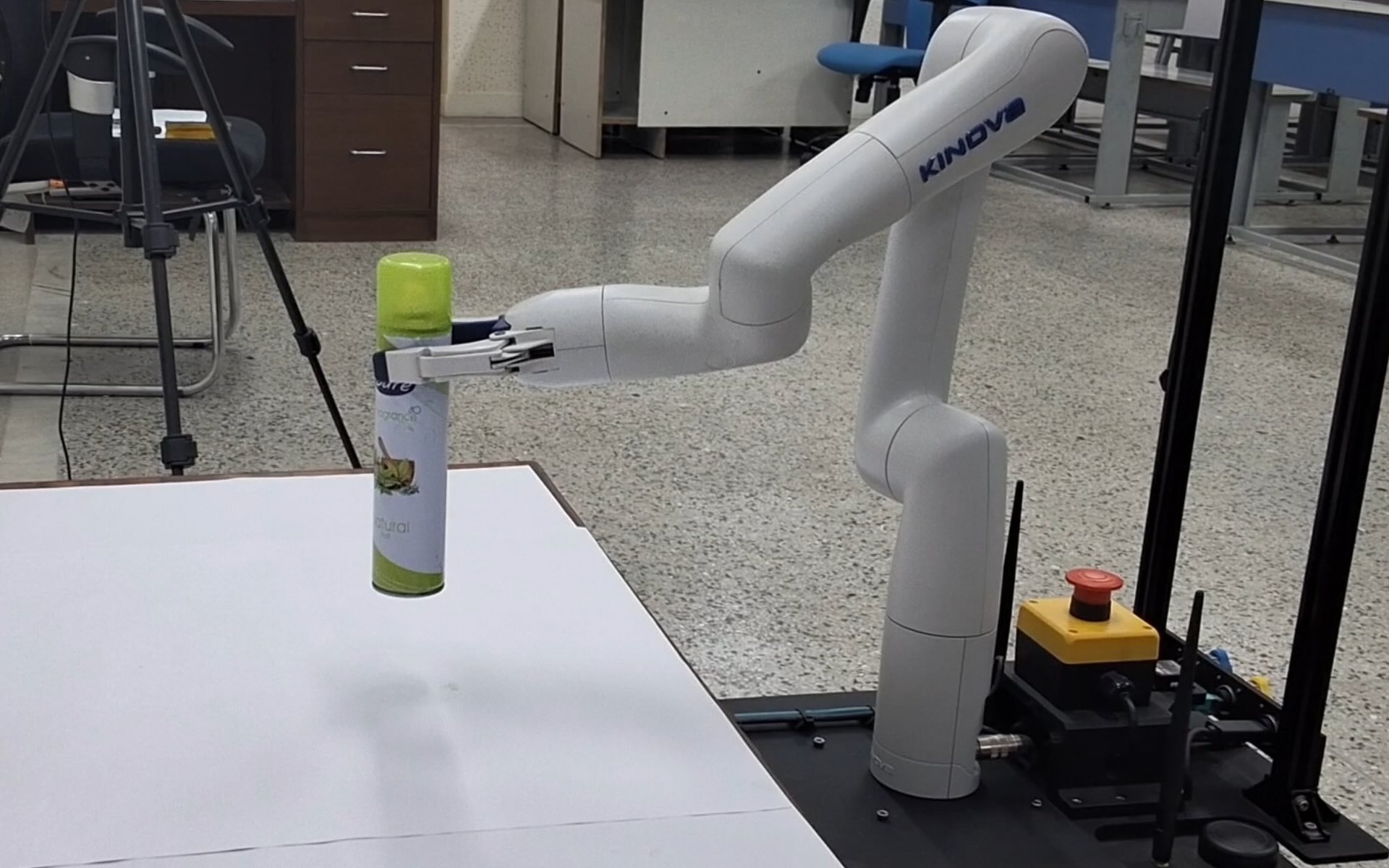}
        \caption{} %
        \label{fig:real_robot_kettle_plan}
    \end{subfigure}
    \begin{subfigure}[t]{0.49\linewidth}
        \includegraphics[width=\textwidth]{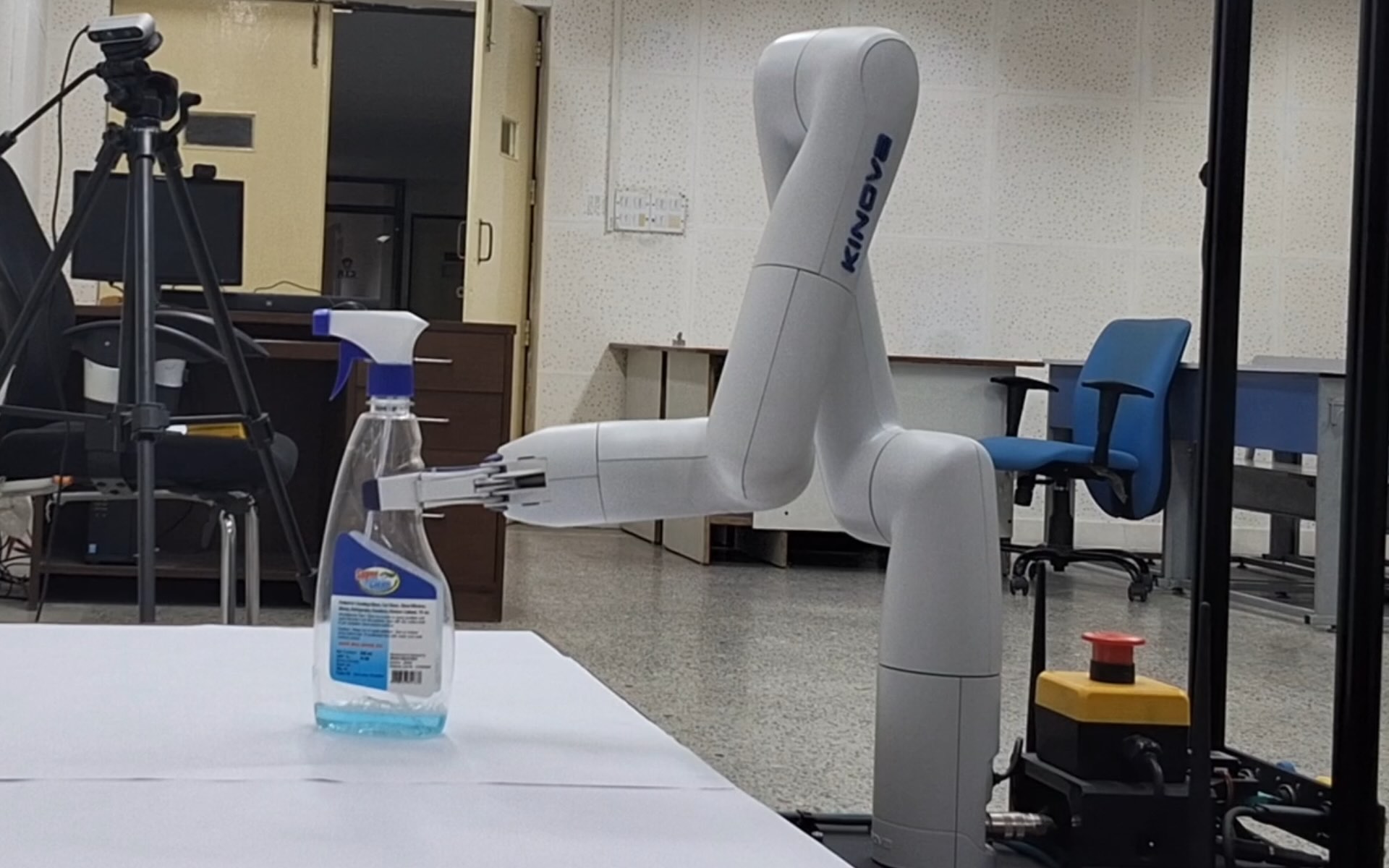}
        \caption{} %
        \label{fig:real_robot_kettle_exec}
    \end{subfigure}
    \caption{
        \textbf{Real-world deployment of GRIM with novel objects.} The system correctly plans and executes task-oriented grasps. The Kinova Gen3 Lite robot successfully executing the planned grasp.
    }
    \label{fig:real_robot}
\end{figure}

To demonstrate the practical applicability of GRIM, we deployed it on a Kinova Gen3 Lite manipulator. The scene is captured by two RGB-D cameras. We used the same 210-instance memory from our simulation experiments, containing no instances of the test objects. We evaluated GRIM on 5 novel objects with associated tasks: a mallet (task:`hammering'), a kettle (task: `pour'), a spray bottle (task: `spray'), an aerosol-can (task: `spray'), and a spoon (task: `scoop'). For each object-task pair, we performed 10 trials. GRIM achieved a high success rate, successfully executing the task-oriented grasp in 39 out of 50 trials. Failures were not due to flawed grasp selection but were instead traced to perception errors; specifically, noise in point cloud reconstruction and calibration inaccuracies were able to disrupt the subsequent 3D alignment stage. Figure \ref{fig:real_robot} shows qualitative examples of successful executions.

\section{Limitations}
As a training-free framework reliant on upstream pre-trained models such as Gemini-Pro, Veo 2, Genie, SAM, and hand-object reconstruction models, it is susceptible to hallucinations, low-quality outputs, or biases inherited from these models, potentially affecting grasp accuracy in edge cases. Additionally, while online inference is efficient at ~10 seconds, the offline memory creation incurs a one-time cost of ~7 minutes per item, which may limit scalability for very large memories. Future work could address these by incorporating robustness checks and expanding real-world benchmarks.

\section{Conclusion}
We have presented GRIM, a training-free framework for task-oriented grasping that demonstrates remarkable generalization capabilities by retrieving and re-aligning functional priors from a diverse memory. Our key innovation is a robust 3D alignment process guided by semantic features, which allows for effective knowledge transfer between objects that are functionally similar but geometrically different. By leveraging generative models and other readily available data sources, GRIM circumvents the data bottleneck that plagues traditional supervised methods. Our extensive experiments show that GRIM significantly outperforms existing training-free and learning-based approaches, particularly in its ability to handle novel objects and tasks.

Future work could explore incorporating explicit geometric reasoning, perhaps through the generation of digital twins \citep{melnik2025digital}, to further refine the alignment and grasp transfer process. Nevertheless, GRIM represents a significant step towards building more general, adaptable, and data-efficient robotic manipulation systems.

\section*{Acknowledgments}
The research reported in this paper was supported by the IHFC-TIH of the Department of Science and Technology, Government of India, Project \#GP/2021/HRI/002, as well as by the German Research Foundation (DFG) through the Collaborative Research Center (Sonderforschungsbereich) 1320, Project-ID 329551904, ``EASE – Everyday Activity Science and Engineering,'' at the University of Bremen (\url{http://www.ease-crc.org/}).

\bibliography{references}

\end{document}